\documentclass[letterpaper, 10 pt, conference]{ieeeconf}  

\IEEEoverridecommandlockouts                              

\usepackage[bookmarks=true]{hyperref}
\usepackage{amsmath,algorithm}
\usepackage{amsfonts}
\usepackage[noend]{algpseudocode}
\usepackage{xcolor}
\usepackage{graphicx}
\usepackage{makecell}
\usepackage{float}
\usepackage{tikz}
\usepackage{pgfplots} 
\usepackage{caption}
\usepackage{subcaption}
\usepackage{multirow}
\usepackage{siunitx}
\usepackage{wrapfig}
\usepackage{siunitx}
\let\labelindent\relax
\usepackage{enumitem}
\usepackage[leftcaption]{sidecap}
\usepackage[font=footnotesize]{caption}
\usetikzlibrary{patterns}
\usetikzlibrary{positioning,shapes.misc}
\tikzset{
    myboxrectangle/.style={rectangle,draw=black,align=center, minimum width=2cm, minimum height=0.8cm},
}
\DeclareMathOperator*{\argmin}{argmin} 

\title{\LARGE \bf
Path-Tree Optimization in Discrete Partially Observable Environments\\ using Rapidly-Exploring Belief-Space Graphs
}

\author{Camille Phiquepal$^{1}$, Andreas Orthey$^{2}$, Nicolas Viennot$^{3}$ and Marc Toussaint$^{2}$
\thanks{$^{1}$Machine Learning \& Robotic Lab, University of Stuttgart, Germany
        {\tt\small camille.phiquepal@ipvs.uni-stuttgart.de}}
\thanks{$^{2}$Learning and Intelligent Systems Lab, TU Berlin, Germany
        {\tt\small toussaint@tu-berlin.de}}
\thanks{$^{3}${\tt\small nicolas.viennot@cs.columbia.edu}}
}

\begin{document}

\maketitle
\thispagestyle{empty}
\pagestyle{empty}

\begin{abstract}
Robots often need to solve path planning problems where essential and discrete aspects of the environment are partially observable.
This introduces a multi-modality, where the robot must be able to observe and infer the state of its environment.
To tackle this problem, we introduce the Path-Tree Optimization (PTO) algorithm which plans a \textit{path-tree} in belief-space. A path-tree is a tree-like motion with branching points where the robot receives an observation leading to a belief-state update. The robot takes different branches depending on the observation received.
The algorithm assumes a deterministic observation model and is composed of three main steps. First, a rapidly-exploring random graph (RRG) on the state space is grown. Second, the RRG is expanded to a belief-space graph by querying the observation model. In a third step, dynamic programming is performed on the belief-space graph to extract a path-tree. The resulting path-tree combines exploration with exploitation i.e.\ it balances the need for gaining knowledge about the environment with the need for reaching the goal. We demonstrate the algorithm capabilities on navigation and mobile manipulation tasks, and show its advantage over a baseline using a task and motion planning approach (TAMP) both in terms of optimality and runtime.
\end{abstract}

\section{Introduction}
Motion planning problems often assume that the environment is fully observable. However, in realistic scenarios, a robot will only have limited access to the environment through its sensors and can therefore only partially observe the state of the world. An example would be a robot, which has a blueprint of a building, but without knowing which doors are open or close. Another example would be a robot in a warehouse, where it has to get a package, but does not know in which location the package resides. In such environments, the robot has a countable number of hypotheses about the environment, which it can observe only in close proximity to certain objects. We say that such environments are partially observable and \emph{multi-modal}.

Multi-modal environments require novel, more general approaches to planning. First, this requires a plan, which takes all possible outcomes into account, such that we have \emph{contingencies} to control the robot irrespective of what observations were made. If a door is closed, the robot should know how to proceed. Second, this requires \emph{optimal motions}, which minimize the expected cost to solve the problem---not only contingencies, but optimal behavior over all modes.

Those requirements are difficult to model with existing frameworks based on task and motion planning (TAMP) or optimization-based methods. Instead, we develop an integrated solution based on belief-space planning, which we call path-tree optimization (PTO). PTO extends previous research work~\cite{tamp-1}\cite{control-tree-1} which introduce tree-like motions computed with optimization based-methods in the respective sub-fields of Task and Motion Planning (TAMP) and Model Predictive Control (MPC). The concept of tree-like motions is extended here to sampling-based path planning, and leverages the strong guarantees of asymptotic optimality to tackle problems challenging for pure optimization-based methods (e.g. due to a high number of local minima). 
In addition, unlike~\cite{tamp-1}, in which the observation actions are planned at the task level, the presented approach incorporates the observation model and the belief-state inference on the motion planning level directly, leading to a unified algorithm. Accordingly, the main contributions of the paper are:
\begin{itemize}[leftmargin=*]
\item a general sampling-based algorithm planning optimal path-trees for partially observable multi-modal problems.
\item the demonstration of the applicability of the method on low-dimensional navigation tasks, as well as on high-dimensional mobile manipulation problems.
\end{itemize}





\section{Related work}
Planning over multi-modal beliefs of the environment has been investigated in the context of active perception. This includes algorithms for SLAM \cite{hsiao2020aras} or active object classification \cite{patten2018monte}. Those methods concentrate primarily on finding good viewpoints, but don't address the motion planning problem. In contrast, with PTO, we concentrate instead on the path planning challenges.

Path planning under partial observability of the environment is related to the broader topic of perception-aware path-planning for which adaptations of classical sampling-based planners~\cite{lavalle1998rapidly}\cite{kavraki1996probabilistic} have been developed.
For example, \cite{hollinger2014sampling}\cite{levine2010information}\cite{dang2020graph} optimize paths on a transition-system grown in a sampling-based fashion. The optimization is w.r.t. an information gathering objective, but does not explicitly model belief-states.

Another group of algorithm considers POMDP-like problems with continuous observations and gaussian belief-states, often to account for localization uncertainty. In particular, our approach shares ideas with BRM~\cite{prentice2010belief}, FIRM~\cite{agha2014firm}\cite{leahy2019control}, or CS-BRM~\cite{zheng2021belief}. The main idea we share is the sampling of a roadmap on which belief-space planning is performed. The main difference lies in the kind of problems we tackle: environment multi-modality in our case vs. continuous uncertainty. This impacts the assumptions made to ensure tractability: countable number of belief states with a deterministic observation model vs. gaussian assumption. From this also follows the different nature of the computed solutions: contingent reactive path-tree vs. paths for \cite{prentice2010belief} or sequences of controllers \cite{agha2014firm}\cite{leahy2019control}\cite{zheng2021belief}.

In the field of object manipulation ConCERRT \cite{pall2018contingent} plans policies reacting to contacts, which reduce the uncertainty over the relative position of the object to manipulate. The contingent-planning aspect is similar to our approach. However, we aim for optimality, whereas \cite{pall2018contingent} does not consider the path costs. The use cases also differ: mobile manipulator with vision based observations vs. table-top object manipulation. In addition, we aim for short planning times (a few seconds) compared to ten minutes or more in~\cite{pall2018contingent}.

Finally, an analogy can be drawn between PTO and Task and Motion Planning (TAMP). In particular, for roadmap-based approaches like \cite{vega2020asymptotically} or \cite{mm-prm}, where a transition system is built across modes. The main difference resides in the meaning of the modes: in PTO modes are belief-states and mode-switches happen when an observation changes the belief-state, whereas modes typically represent different tasks in TAMP. The difference in planning problems also lead to the optimization of contingent path-trees, optimized over all modes simultaneously, instead of paths.

\section{Problem formulation}
\begin{figure}
	\begin{subfigure}[b]{0.5\linewidth}
		\centering
		\includegraphics[width=\textwidth]{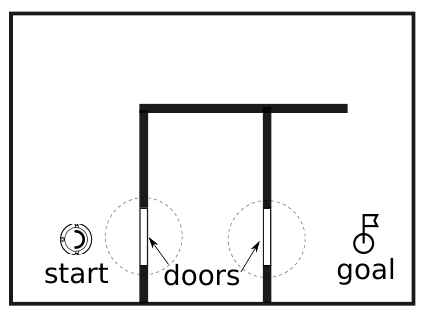}
		  \caption{Example of planning problem}
         \label{fig:problem}
	\end{subfigure}\hfill
	\begin{subfigure}[b]{0.5\linewidth}
		\centering
		\includegraphics[width=\textwidth]{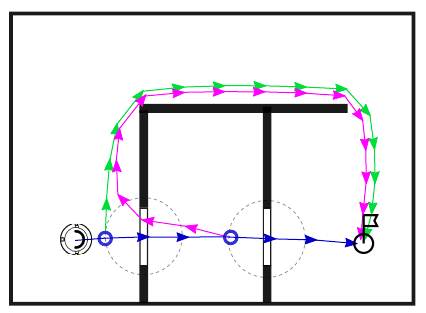}
		\caption{Reactive Path-tree}
    \label{fig:path_tree}
	\end{subfigure}
	\caption{The robot has to reach the goal region. The doors state (open vs. closed) is only observable in their vicinity. PTO plans path-trees with observation points (blue circles) and path-contingencies (blue, green, magenta) reacting to the observations.}
	\label{fig:intro}
\end{figure}

We optimize policies in a context of mixed-observability. The robot state is fully observable, but discrete and essential parts of the environment are only partially observable. In the example of the Fig.~\ref{fig:problem} this models the fact that each door can be open or closed.
 
To capture this mixed observability, we adopt a compound state representation where a state is composed of 2 parts:
\begin{itemize}
\item $x \in \mathbb{R}^n$ is the robot state from a continuous space, it is fully observable.
\item $s \in \mathcal{H}$ is a discrete state from a finite set of world hypotheses $\mathcal{H}$.
\end{itemize}

\begin{figure}
 \center{\includegraphics[width=0.45\textwidth]{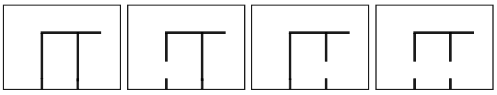}}
 \caption{Partially observable discrete state: With 2 partially observable doors, there are 4 possible states of the environment.}
 \label{fig:multiple_worlds}
\end{figure}

In the problem introduced in Fig.~\ref{fig:problem} the variable $s$ can take four possible values corresponding to the possible states of the environment as illustrated on Fig.~\ref{fig:multiple_worlds}.

The robot is not oblivious about the likelihood of each state hypothesis. On the problem introduced on Fig.~\ref{fig:problem}, when the robot is in the vicinity of a door (symbolized with the circles), it receives an observation indicating whether the door is open or not.

Planning is performed in belief-space, where a belief-state $b$ is a probability distribution over the possible states.

\subsection{Path-tree}
The algorithm consists in planning a path-tree. A path-tree consists of paths that branch into multiple possible paths where the robot receives an observation leading to a belief-state update. As schematically shown on Fig.~\ref{fig:path_tree}, the path-tree starts from a single root node and finishes with leaf nodes satisfying the goal condition. The different branches of the path-tree are the different planned contingencies. To be complete, a path-tree must end inside the goal region for all $s \in \mathcal{H}$.

In addition, we assume the observation model to be binary, meaning that $p(o|x, s) \in \{0.0, 1.0\}, x \in \mathbb{R}^n, \forall s \in \mathcal{H}, \forall o \in O$, where $o$ is an observation from the observation space $O$. This property is important. It guarantees that the number of belief-states is finite and can be enumerated which is needed for
the graph expansion to belief-state presented in \ref{section:graph_expansion_belief_space}. On Fig.~\ref{fig:problem}, it means that observations indicate whether the door is open or not without uncertainty. In addition, it typically implies that the number of branchings $N_o$ stays small compared to the total number of states $N$ on the path-tree. In other words $N_o \ll N$. It can be understood easily on the example: once an observation has been received for a door, the agent knows with certainty if the door is open or not, such that the next observations of the same door do not lead to an update of the belief-state. 

\subsection{Optimization objective}
We note $\psi$ a path-tree, and $(u, v)$ consecutive nodes on the tree $\psi$. Nodes are associated with a robot configuration and a belief state. The motion cost between two nodes $u$ and $v$ is given by a cost function $C(u, v)$. In addition, we note $p(u | \psi, b_0)$ the probability to reach a node $u$. It depends on the initial belief-state $b_0$ and the observation model. 

In the presence of uncertainty we minimize the expectation of the motion costs, such that the partially observable multi-modal optimization problem is defined as follows.
\begin{subequations}
\label{eq:optimization_objective}
\begin{flalign}
\psi^*&=\argmin_{\psi} \sum_{(u, v) \in \psi} C(u, v) p(v | \psi, b_0),&& \label{eq:cost_min}\\
&\text{s.t.} \notag \\
&\forall s \in \mathcal{H}\ \exists\ l \in \mathcal{L}(\psi) \ |\ G(l), \label{eq:goal_constraint}\\
&\mathcal{V}(u, v), \ \forall (u, v) \in \psi \label{eq:validity}, \\
& b_v(s) = \frac{p(o | s) b_u(s)}{\sum\limits_{s^{\prime}} p(o | s^{\prime}) b_u(s^{\prime}) }, \forall s \in \mathcal{H}, \forall (u,o,v) \in \psi \label{eq:belief_dynamic}
\end{flalign} 
\end{subequations}
where $\mathcal{L}(\psi)$ gives the leaf nodes of $\psi$, and $G(l)$ is the goal predicate, indicating whether a node fulfills the goal conditions. Eq.~\eqref{eq:goal_constraint} states that $\psi$ shall be complete i.e. there is a leaf node satisfying the goal condition for each state $s$.

Eq.~\eqref{eq:validity} expresses that $\psi$ shall be composed of feasible motions (e.g. collision free, feasible given the robot motion model), and $\mathcal{V}(u, v)$ is the predicate encoding the validity of the transition between $u$ and $v$.

Finally, Eq.~\eqref{eq:belief_dynamic} corresponds to the Bayesian updates of the belief state. The symbol $o$ is the observation received when transitioning from $u$ to $v$. The belief-state of the node $v$ is noted $b_v$, and $p(o|s)$ is the observation model.

The problem formulation does not contain any term explicitly incentivizing the robot to explore its environment. The balance exploration / exploitation will emerge as a result of the minimization of the expected motion costs (see \ref{section:policy-extraction}).

\section{Path-tree optimization (PTO)}
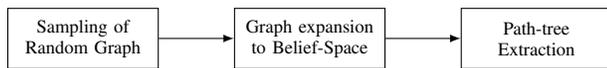
\begin{figure}[!htb]
\scriptsize
\begin{tikzpicture}[>=latex] 
\node[myboxrectangle] (RRG) {Sampling of\\ Random Graph};
\node[myboxrectangle] (BS) [right =of RRG] {Graph expansion \\to Belief-Space} edge [<-] (RRG);
\node[myboxrectangle] (PO) [right =of BS] {Path-tree\\Extraction} edge [<-] (BS);
\end{tikzpicture}
 \caption{PTO Algorithm overview. It contains three main steps corresponding to the algorithms described in the sections \ref{section:rapidly-exploring-random-graph}, \ref{section:graph_expansion_belief_space} and \ref{section:policy-extraction}.}
 \label{algorithm_overview}
\end{figure}

To build a path-tree satisfying the specification of Equation \ref{eq:optimization_objective}, we proceed stepwise. First a transition system is grown in a sampling-based fashion until the existence of a solution is guaranteed. This corresponds to the first two steps schematized on Fig. \ref{algorithm_overview}. In a second step, the optimal path-tree is extracted using dynamic programming. 

\subsection{Interface between the algorithm and the application layer}

The connection between the core of the algorithm and a given planning problem is achieved via 4 functions that the application layer provides: 
\begin{itemize}
\item \textsc{StateCheck}, which takes a robot configuration $x$ as input and returns the list of worlds in which the configuration is valid.
\item \textsc{TransitionCheck}, which takes two robot configurations as inputs and returns the list of worlds in which the robot configurations are valid.
\item \textsc{GoalCheck}, which takes a robot configuration as input and returns the list of worlds in which the robot configuration fulfills the goal conditions.
\item \textsc{Observe}, which receives a robot configuration and a belief-state as inputs and returns the possible output belief-states.
\end{itemize}

\textsc{StateCheck}, \textsc{TransitionCheck} and \textsc{GoalCheck} are akin to the functions required by standard path-planning frameworks like OMPL~\cite{sucan2012the-open-motion-planning-library}, but they are more general: the return type is a list of worlds instead of a boolean.

The \textsc{Observe} function is directly linked to partial observability and is specific to this algorithm. It allows belief-state inference. Going back to the example of Fig. \ref{fig:problem}, \textsc{Observe} returns an unchanged belief-state for all robot configurations that are outside of the door visibility zone, because no observation can be received, and therefore no belief-state inference is done. On the other hand, up to two possible belief-states are returned for robot configurations inside the door visibility zone. They correspond to the updated belief-states for the two possible observations.

The next sections detail how the calls to those functions are orchestrated by the algorithm to build a path-tree.

\subsection{Rapidly-exploring Random Graph}
\label{section:rapidly-exploring-random-graph}

In this first step, a random-graph is grown in a sampling-based fashion. To avoid the curse of dimensionality, sampling is not performed in belief-space directly but in the robot configuration space. The random graph is an intermediate representation and will be expanded to belief-state in a second step as described in the next section.

The nodes of the random graph are associated with:
\begin{itemize}
\item a robot configuration $x$, which is randomly sampled.
\item a list of worlds $\mathcal{R}$ in which the node is guaranteed to be reachable. It is obtained by composing the reachability of the node's parents with the results of the \textsc{TransitionCheck}.
\item a list of worlds $\mathcal{F}$ in which the robot configuration is fulfilling the goal condition. This is obtained by querying the function \textsc{GoalCheck}. 
\end{itemize} 

The edges of the random graph are associated with:
\begin{itemize}
\item a list of worlds $\mathcal{W}$ in which the transition is valid. This is obtained by calling the function \textsc{TransitionCheck}.
\end{itemize}



\begin{SCfigure}
 \caption{Random Graph expansion: For a new sample \textit{a)}, the new state is computed steering from the closest node having a reachability $\mathcal{R}$ compatible with the sampled world \textit{w}, see \textit{b)} and \textit{c}).}
 \includegraphics[width=0.275\textwidth]{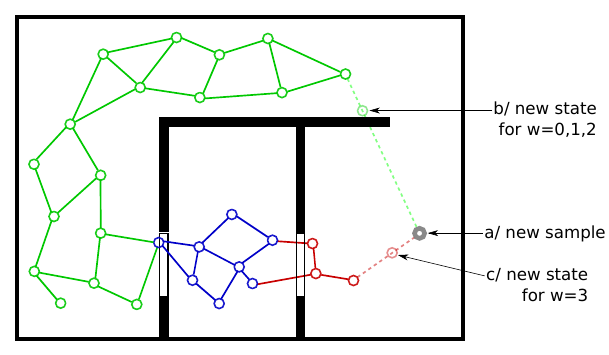}
 \label{fig:expansion}
\end{SCfigure}

The random-graph creation is described in Alg.~\ref{alg:rrg}. First, the state is sampled (line 4). Then the new state is steered from a neighbor node of the graph (line 7). Unlike RRT where the new state is steered from the nearest neighbor, here, the selected neighbor is the nearest neighbor having world reachabilities $\mathcal{R}$ containing a uniformly sampled world $w$ (line 5), as illustrated on Fig.~\ref{fig:expansion}. This additional condition for the neighbor selection is to ensure that the random graph contains paths to the goal for each world.
 


If the new state is valid for at least one world (line 9), the goal conditions are checked (line 10) and a new node is added to the random-graph (line 11).
Finally, the new node is connected to nodes within an adaptive radius $r$~\cite{solovey2020revisiting}, and having world validities containing $w$ (lines 13 to 16).

This procedure is repeated at least until the graph is complete, meaning that it will allow the successful extraction of a solution path-tree (see line 3). To implement the function \textsc{IsComplete} we assume in our examples that the transitions are \textit{symmetrical}, i.e. if a motion exists from a node $u$ to $v$, then a motion between $v$ and $u$ also exists. Under this assumption, the random-graph contains a solution path-tree as soon as the set of leaf nodes is complete, meaning that it covers every possible world, i.e. $\bigcup_{u \in \mathcal{\mathcal{L}(G)}} \mathcal{F}_u \cap \mathcal{R}_u = \mathcal{H}$. Indeed, the fact that the set of leaf nodes is complete implies that for each possible world, a path from the root to a leaf exists. A naive solution is therefore to execute those paths in sequence, and  potentially \textit{backtrack} to the root node if a path doesn't reach the goal (hence the required assumption regarding the symmetrical transition). In practice, once the completeness condition is reached, a solution can be extracted which is more optimal than the aforementioned worst-case backtracking strategy. A minimum number of iterations is also specified, to eventually expand the random-graph beyond the completeness threshold to improve the quality of the path-trees.

\begin{algorithm}
\caption{Rapidly-exploring Random Graph}
\label{alg:rrg}
\begin{algorithmic}[1]
\Function{BuildRRG}{$q_{start}$, $i_{min}$}
	\State $\mathcal{G}$.\Call{init}{$q_{start}$}; $i \gets 0$
	\While{$\neg$ \Call{IsComplete}{$\mathcal{G}$} \textbf{or} $i < i_{min}$}
    	\State $q_{rand} \gets$ \Call{SampleState()}{}
    	\State $w \gets$
        \Call{SampleWorld()}{}
    	\State $q_{near} \gets$ \Call{Nearest}{$q_{rand}, w$}
    	\State $q_{new} \gets$ \Call{Steer}{$q_{near}, q_{rand}$}
    	\State \textcolor{cyan}{\footnotesize/*get relevant info of the new state and add it to the graph*/}
    	\If{\Call{StateCheck}{$q_{new}$} $\neq \emptyset$}
    	    \State $\mathcal{F} \gets$ \Call{GoalCheck}{$q_{new}$}
    		\State $\mathcal{G}$.\Call{AddNode}{$q_{new}, \mathcal{F}$}
    		\State \textcolor{cyan}{\footnotesize/*get relevant edges info and add them to the graph*/}
    		\For{$q_{near} \in$ \Call{Nearests}{$q_{new}, w, r$}}
    			\State $\mathcal{W} \gets$ \Call{TransitionCheck}{$q_{near}, q_{new}$}
    			\If{$\mathcal{W} \neq \emptyset$}
    				\State $\mathcal{G}$.\Call{AddEdge}{$q_{near}, q_{new}, \mathcal{W}$}
    			\EndIf
    		\EndFor
    	\EndIf
    	\State $i \gets i+1$
    \EndWhile
\EndFunction
\Statex
\end{algorithmic}
\end{algorithm}
Each node is kept connected to multiple neighbors because the best parent at this stage is ambiguous: it is only defined for a given belief-state. The optimal traversal will be determined on the belief-space graph (see \ref{section:policy-extraction}).
\subsection{Graph expansion to belief-space}
\label{section:graph_expansion_belief_space}
\begin{figure}
 \center{\includegraphics[width=0.45\textwidth]{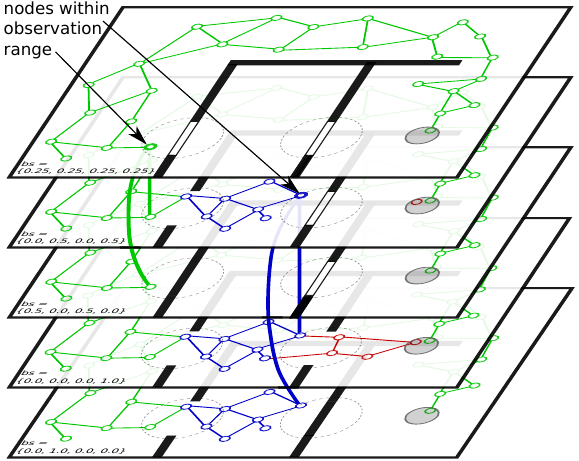}}
 \caption{Random Graph expanded to belief-state: The belief-states are represented as \textit{layers} of the belief-graph. The nodes within the observation range potentially lead to a belief-state update and therefore have edges transitioning to other belief-states (thick vertical edges).}
 \label{fig:belief_layers}
\end{figure}

The random graph $\mathcal{G}$ is an intermediate representation that cannot be used directly to optimize path-trees. In this step, a transition system in belief-space, or belief-graph $\mathcal{B}$ is constructed out of $\mathcal{G}$ by querying the observation model. 
The nodes of the belief-graph are associated with:
\begin{itemize}
\item a robot-configuration $x$.
\item a belief-state $b$.
\end{itemize}
The edges of the belief-graph are associated with:
\begin{itemize}
\item the observation $o$ making the transition between the beliefs on the incoming and outcoming nodes.
\end{itemize}
The belief-graph can be understood as the random-graph replicated over several layers, where each layer is a different belief-state, as shown on Fig.~\ref{fig:belief_layers}. The transitions within one layer correspond to robot motions, whereas the transitions from one layer to another correspond to the integration of an observation leading to a belief-state update. The motion transitions are, in general, not identical across belief-states. For example, on Fig.~\ref{fig:belief_layers}, the transitions crossing through the doors exist only in beliefs compatible with an open door.
  
The construction procedure is given by the Alg.~\ref{alg:belief}. First, the edges of $\mathcal{G}$ are replicated to the belief-states compatible with the edge's valid worlds (line 3 to 7).

Second, the observation model is called via the function \textsc{OBSERVE} to identify where edges should be created between belief-states (lines 9 to 12). The belief-state-transition edges, due to observations, are represented by vertical lines going from one layer to another on Fig.~\ref{fig:belief_layers}.

\begin{algorithm}
\caption{Creation of the belief-graph}
\label{alg:belief}
\begin{algorithmic}[1]
\Function{BuildBeliefGraph}{$b_{start}, \mathcal{G}$}
	\State \textcolor{cyan}{\footnotesize/*connect nodes within the same belief*/}
	\For{$e \in$ $\mathcal{G}.edges$}
		\For{$b \in$ \Call{Beliefs}{$e.world\_validities$}}
			\State $u \gets \mathcal{B}$.\Call{AddNode}{$b, e.from$}
			\State $v \gets \mathcal{B}$.\Call{AddNode}{$b, e.to$}
			\State $\mathcal{B}$.\Call{AddEdge}{$u, v$}
		\EndFor
	\EndFor
	\State \textcolor{cyan}{\footnotesize/*create transitions between beliefs due to observations*/}
	\For{$u \in$ $\mathcal{B}.nodes$}
	   	\State $V \gets$ \Call{Observe}{$u$}
		\For {$v \in V$}
			\State $\mathcal{B}$.\Call{AddEdge}{$u$, $v$}
		\EndFor
	\EndFor
\EndFunction
\end{algorithmic}
\end{algorithm}

The belief-graph $\mathcal{B}$ is significantly larger than the random graph $\mathcal{G}$ since there are multiple belief-states corresponding to the same robot configuration. However, its construction only calls the observation model (function \textsc{Observe}). In contrast, the random-graph is smaller but its creation calls the expensive collision checks (\textsc{StateCheck} and \textsc{TransitionCheck}) in addition to the nearest neighbor search.

\subsection{Policy extraction}
\label{section:policy-extraction}

Now that $\mathcal{B}$ is built and contains at least one path-tree satisfying the goal conditions (Eq.~\ref{eq:goal_constraint}), the goal is to find the optimal path-tree by minimizing the expected costs (Eq.~\ref{eq:cost_min}).

The expected costs to goal are computed using dynamic programming by iteratively applying Bellman updates on each node of $\mathcal{B}$, as described by the Algorithm~\ref{alg:expected_costs}.

The algorithm is similar to the Dijkstra algorithm \cite{Sniedovich2006DijkstrasAR} in the way the nodes are prioritized using a priority queue (lines 2 to 13). For edges corresponding to a robot motion, the Bellman update is also the same as in Dijkstra (lines 15 and 16). However, for edges corresponding to an observation, the Bellman update of the parent node differs: it is the sum of the children's expected costs weighted by their respective branching probabilities (lines 17 to 19).
\begin{algorithm}
\caption{Computation of the expected costs to goal}
\label{alg:expected_costs}
\begin{algorithmic}[1]
\Function{ComputeExpectedCostToGoal}{$\mathcal{B}$}
	\State $C \gets $ \Call{Array()}{} \textcolor{cyan}{\footnotesize// Expected costs to goal indexed by node id}
	\State $Q \gets $ \Call{PriorityQueue()}{}
	\State \textcolor{cyan}{\footnotesize/*Initialization*/}
	\For{$n \in$ $\mathcal{B}.nodes$}
		\If{\Call{IsFinal}{$n$}}
			\State $C[n] \gets 0.0$
			\State $Q$.\Call{Push}{$n, 0.0$}
		\Else
			\State $C[n] \gets +\infty$
		\EndIf
	\EndFor
	\State \textcolor{cyan}{\footnotesize/*The main loop*/}
	\While{$\neg$ \Call{IsEmpty}{$Q$}}
		\State $v \gets$ \Call{Pop}{$Q$}
		\For {$u \in$ \Call{Parents}{$v$}}
			\State \textcolor{cyan}{\footnotesize/*Bellman update dependent on the edge type*/}
			\If{\Call{IsActionEdge}{$u, v$}}
				\State $c \gets$ \Call{Cost}{$u$, $v$}$+ C[v]$
			\ElsIf{\Call{IsObservationEdge}{$u, v$}}
				\State $\mathcal{W} \gets $ \Call{ObservationChildren}{$u$}
				\State $c \gets \sum_{\nu \in \mathcal{W}}{ p(\nu | u) \times C[\nu]} $
			\EndIf
			\If{$c < C[u]$}
				\State $C[u] \gets c$
				\State $Q$.\Call{Push}{$u, c$}
			\EndIf
		\EndFor
	\EndWhile
\EndFunction
\end{algorithmic}
\end{algorithm}

Once the expected costs to goal are known, the optimal path-tree $\psi^*$ can be built straightforwardly starting from the root and recursively appending the best next child, or next best children (on nodes with observation branching).

It is noteworthy, that the expected costs are computed for all nodes of $\mathcal{B}$, such that a path-tree can be extracted from each node (not only from the root node). In practice, this can be used for quick re-planning at execution time. 


Optimizing the expected costs to goal naturally results in an exploration vs. exploitation trade-off, i.e. path-trees balance the need to move towards configurations providing informative observations vs. advancing towards the goal.

\subsection{Path-tree refinement}
\label{section:refinement}

Despite asymptotic optimality guarantee, a path-tree may contain unnatural or jerky motions due to the random nature of the random-graph creation. While it is possible to increase the number of iterations, this can become costly in terms of runtime. To tackle this, we add a refinement step, working piecewise: path-pieces between observation branchings are refined using the partial-shortcut method \cite{geraerts2007creating}. Observation nodes connecting path-pieces are not modified by the refinement procedure. 

\subsection{Completeness and optimality}

We argue that PTO is probabilistically complete and asymptotically optimal, i.e. it will eventually solve Eq.~\eqref{eq:optimization_objective} to the optimal solution. This is true under the assumption that our observation model is well behaved and does not introduce zero-measure configurations changing the belief state. Our reasoning is divided into two steps. First, since we use an asymptotically optimal sampling-method~\cite{karaman2011sampling, solovey2020revisiting} on every modality (Alg.~\ref{alg:rrg}), together with a uniform modality sampler, PTO will eventually converge to the optimal solution on each mode. Second, by inter-connecting edges between modalities, we guarantee that we account for possible mode-changes which can only happen in observation nodes (Alg.~\ref{alg:belief}). The final belief-graph will therefore contain the optimal solution, which we can then extract using the Dijkstra-like algorithm in our policy extraction step (Alg.~\ref{alg:expected_costs}). This combination guarantees that the optimal solution will eventually be attained in the limit.

\section{Experiments}

PTO is implemented in the Rust programming language \cite{matsakis2014rust}. The application layer is implemented in C++ using MoveIt \cite{coleman2014reducing}. The source code is available for reference \footnotemark \footnotetext{\href{https://github.com/cambyse/po-rrt}{https://github.com/cambyse/po-rrt}}.

\subsection{Mobile robot navigation} \label{experiment:navigation}
We consider two planning problems similar to the initial example introduced in Fig.~\ref{fig:intro}. The robot is the Kobuki platform. The robot has to reach a predefined goal region, but at planning time, it doesn't know which doors are open. The observation model simulates that the sensor mounted on the Kobuki and the perception pipeline allow the robot to know if a door is open once the robot is close to a door (less than 1.5 m) and has a non-occluded line of sight to the door. We plan for the two-dimensional position of the robot.

\begin{figure}[h]
	\begin{subfigure}[b]{0.485\linewidth}
		\centering
		\includegraphics[width=\textwidth]{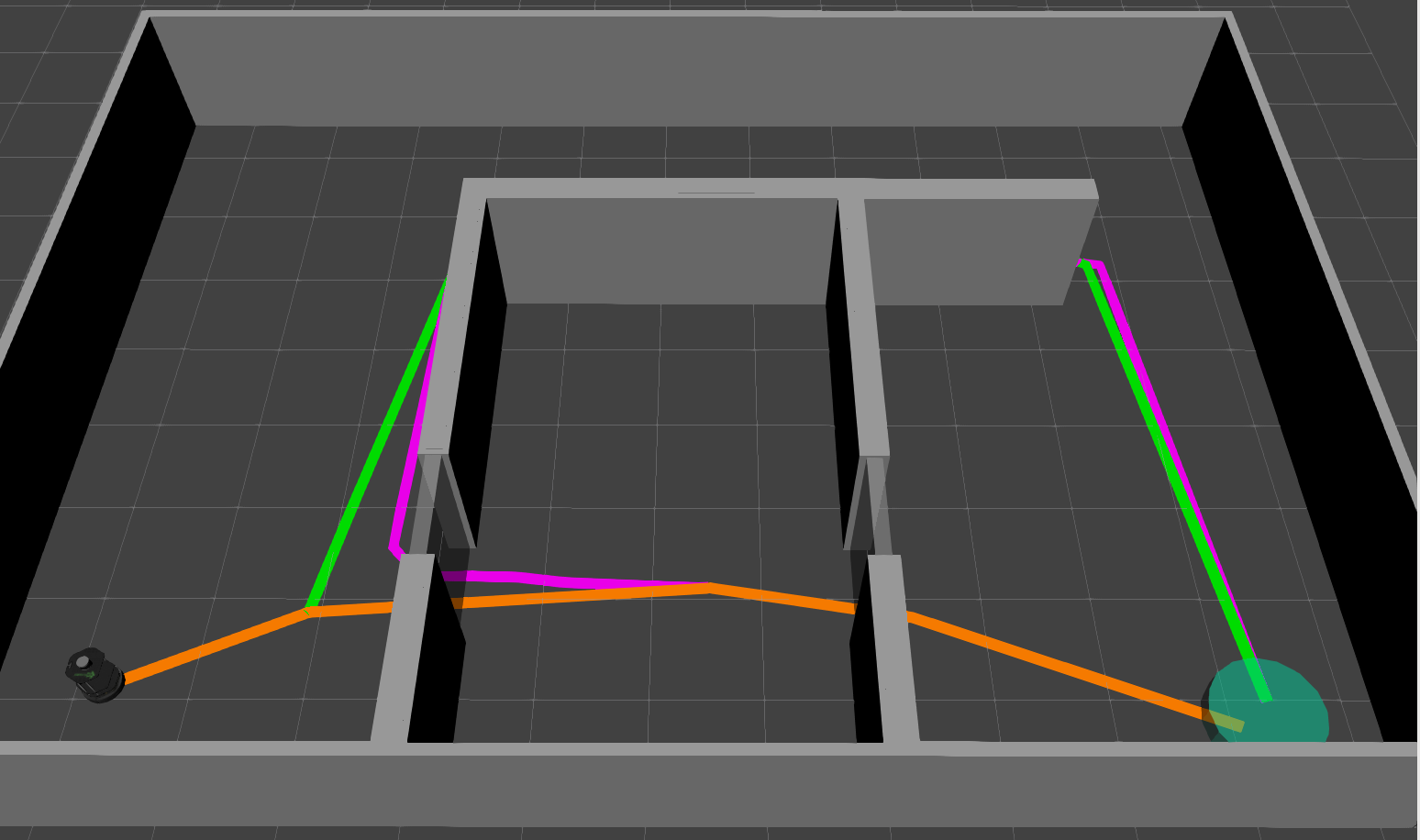}
		 \caption{Path-tree obtained with a high likelihood that doors are open (80\%). The path-tree has 2 branching points corresponding to the observation of the 2 doors.}
         \label{fig:0p7_belief_state}
	\end{subfigure}\hfill
	\begin{subfigure}[b]{0.485\linewidth}
		\centering
		\includegraphics[width=\textwidth]{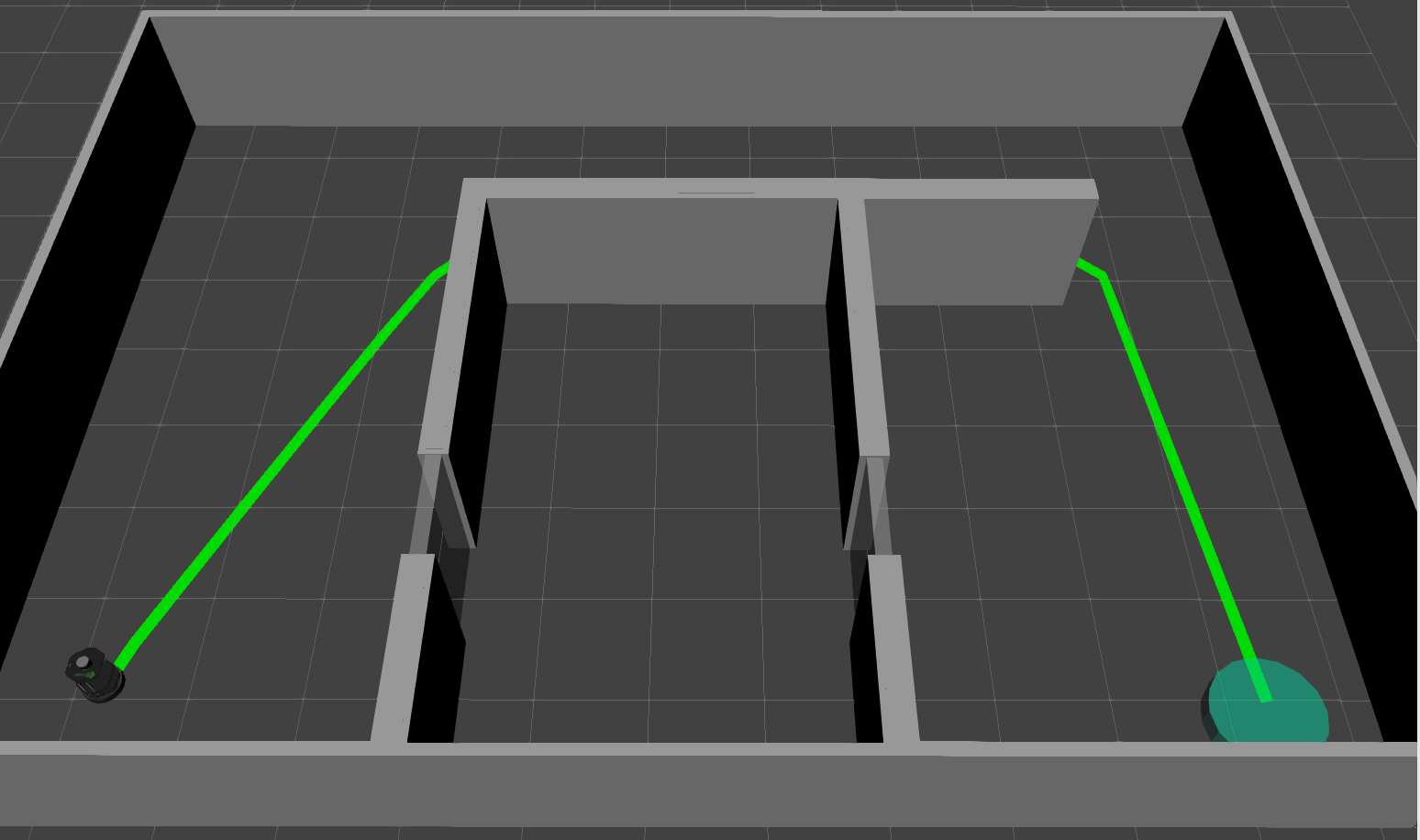}
		\caption{Path-tree obtained with 50\% of likelihood that doors are open: It is not advantageous to attempt the direct way through the doors.}
    \label{fig:0p25_belief_state}
	\end{subfigure}
	\caption{Problem-A: The robot has to reach the green region on the right.}
	\label{fig:problem_a_influence_belief_states}
\end{figure}

\begin{wrapfigure}{l}{0.30\textwidth}\vspace{-10pt}
\includegraphics[width=0.30\textwidth]{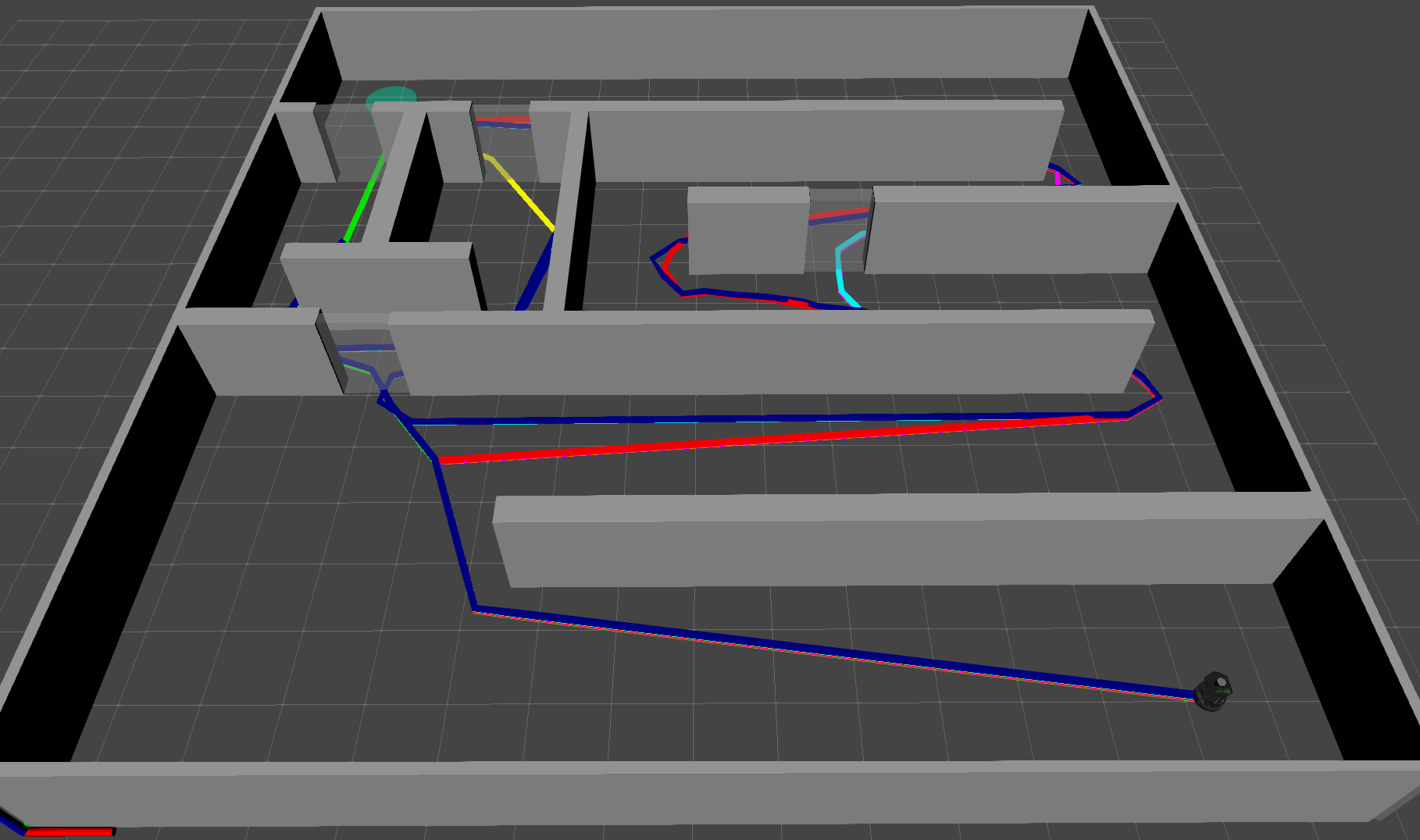}
\caption{Path-tree for Problem-B: The map has four doors leading to a more complex path-tree.}
\label{fig:problem_b}
\vspace{-10pt}
\end{wrapfigure}

The first problem, called problem-A (Fig.~\ref{fig:problem_a_influence_belief_states}) has two uncertain doors. The second problem, called problem-B (Fig.~\ref{fig:problem_b}) is more complex, the map is larger, there are four doors, and more obstacles.


In this example, we see the strong influence of the initial belief-state on the topology of the path-tree: Fig.~\ref{fig:0p25_belief_state} shows the path-tree obtained with an initial probability of only 50\% that the doors are open. In that case, it is not worthy to attempt the way through the doors, the planned path takes the longer, but sure way to the goal. The path-tree boils down to a sequential path without observing branching points.

Fig.~\ref{fig:0p7_belief_state} shows the path-tree obtained with a higher likelihood that each door is opened (80\%). In that case, it is advantageous to attempt the direct path through the doors. The path-tree has two branching points  corresponding to the observations of the two doors. 

\begin{figure}[h]
	\begin{subfigure}[b]{0.33\linewidth}
		\centering
		\includegraphics[width=\textwidth]{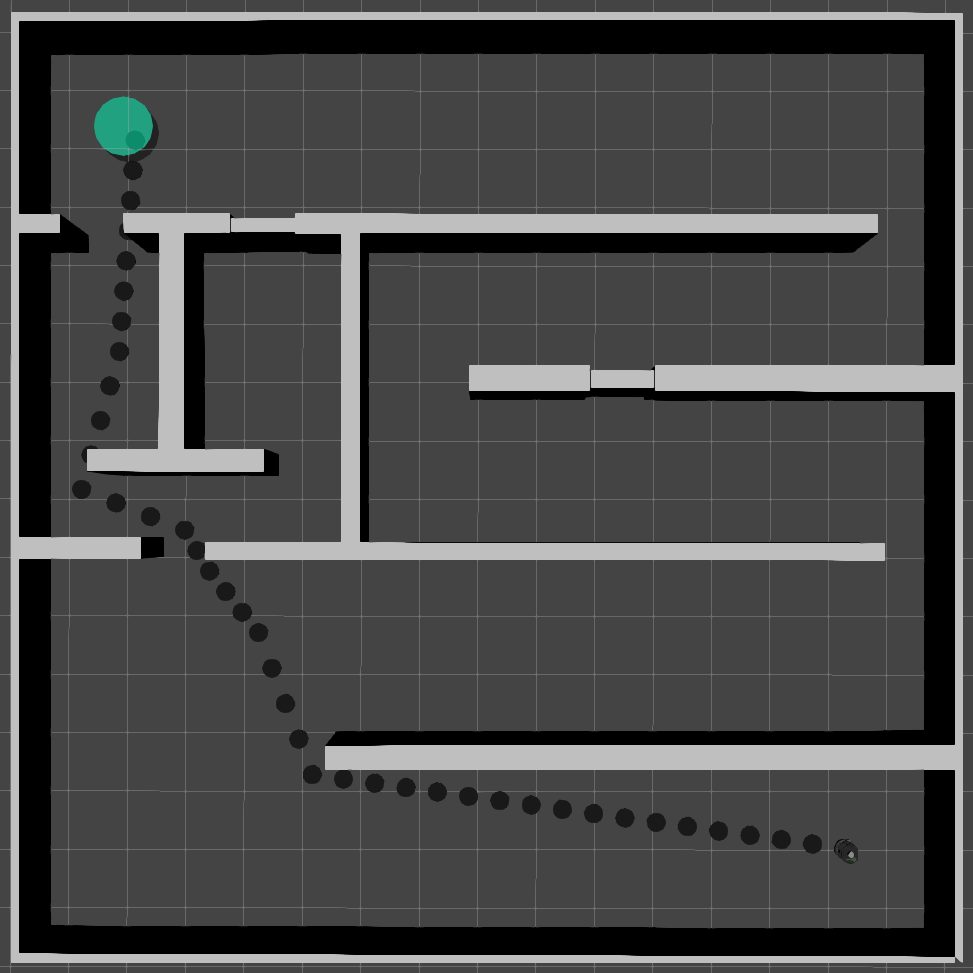}
         \label{fig:problem_b_world_0}
	\end{subfigure}\hfill
	\begin{subfigure}[b]{0.33\linewidth}
		\centering
		\includegraphics[width=\textwidth]{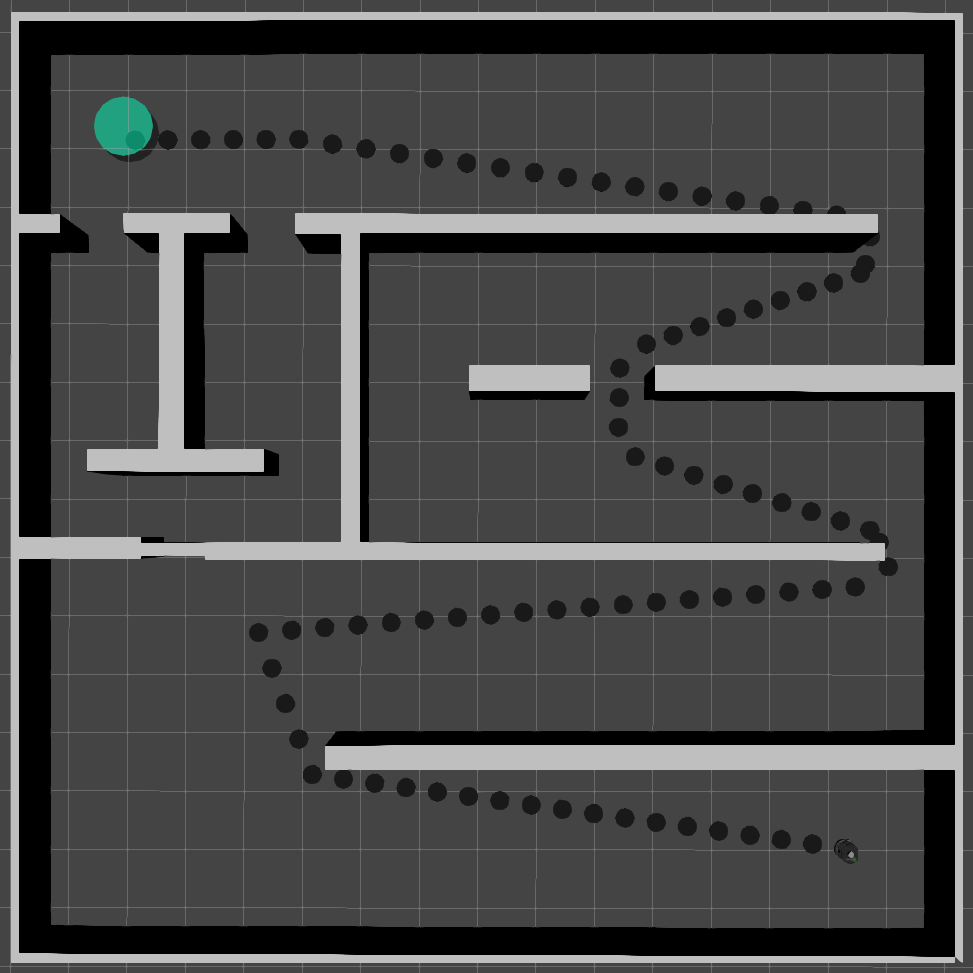}
    \label{fig:problem_b_world_1}
    \end{subfigure}\hfill
    \begin{subfigure}[b]{0.33\linewidth}
		\centering
		\includegraphics[width=\textwidth]{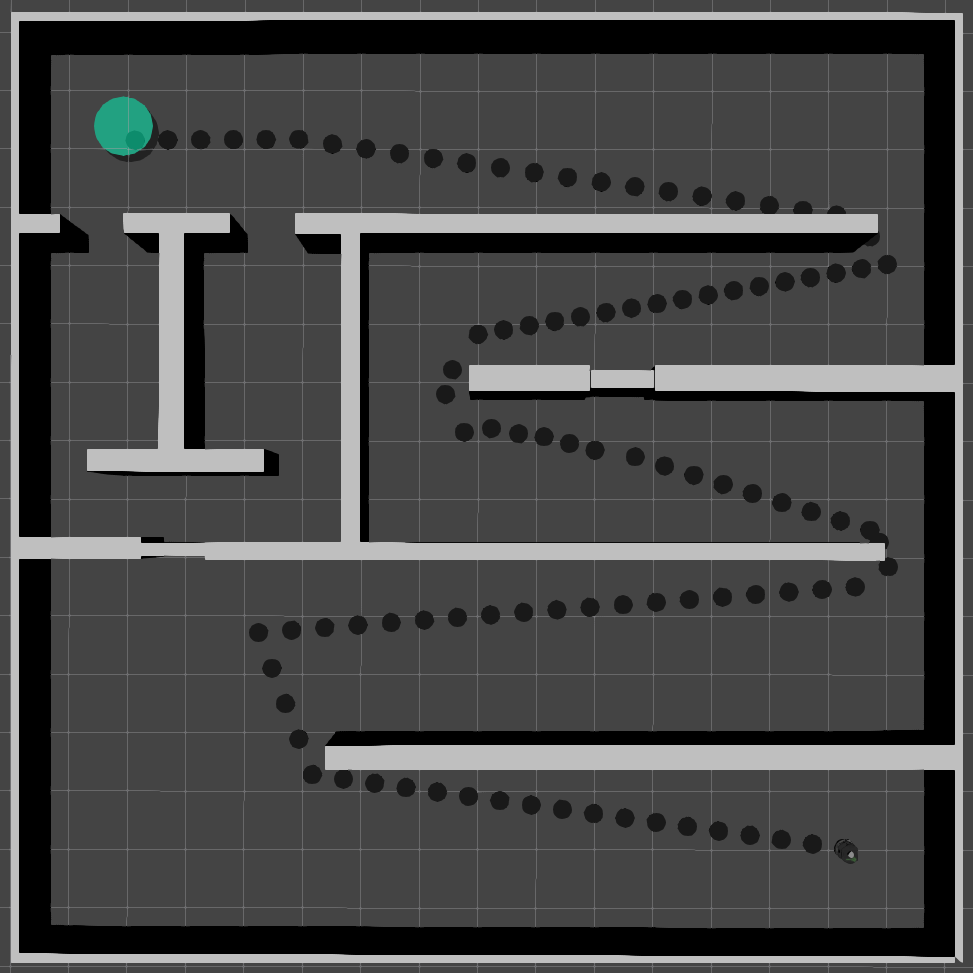}
         \label{fig:problem_b_world_2}
	\end{subfigure}\hfill
	\begin{subfigure}[b]{0.33\linewidth}
		\centering
		\includegraphics[width=\textwidth]{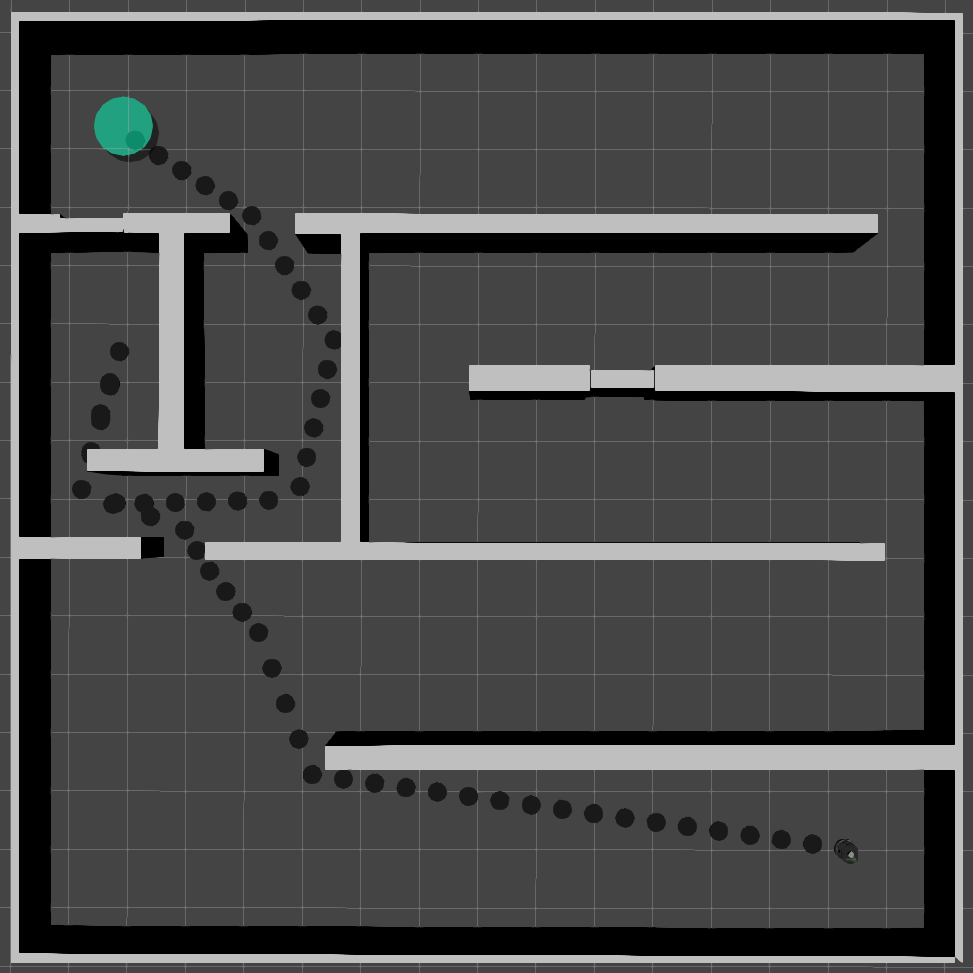}
         \label{fig:problem_b_world_3}
	\end{subfigure}\hfill
	\begin{subfigure}[b]{0.33\linewidth}
		\centering
		\includegraphics[width=\textwidth]{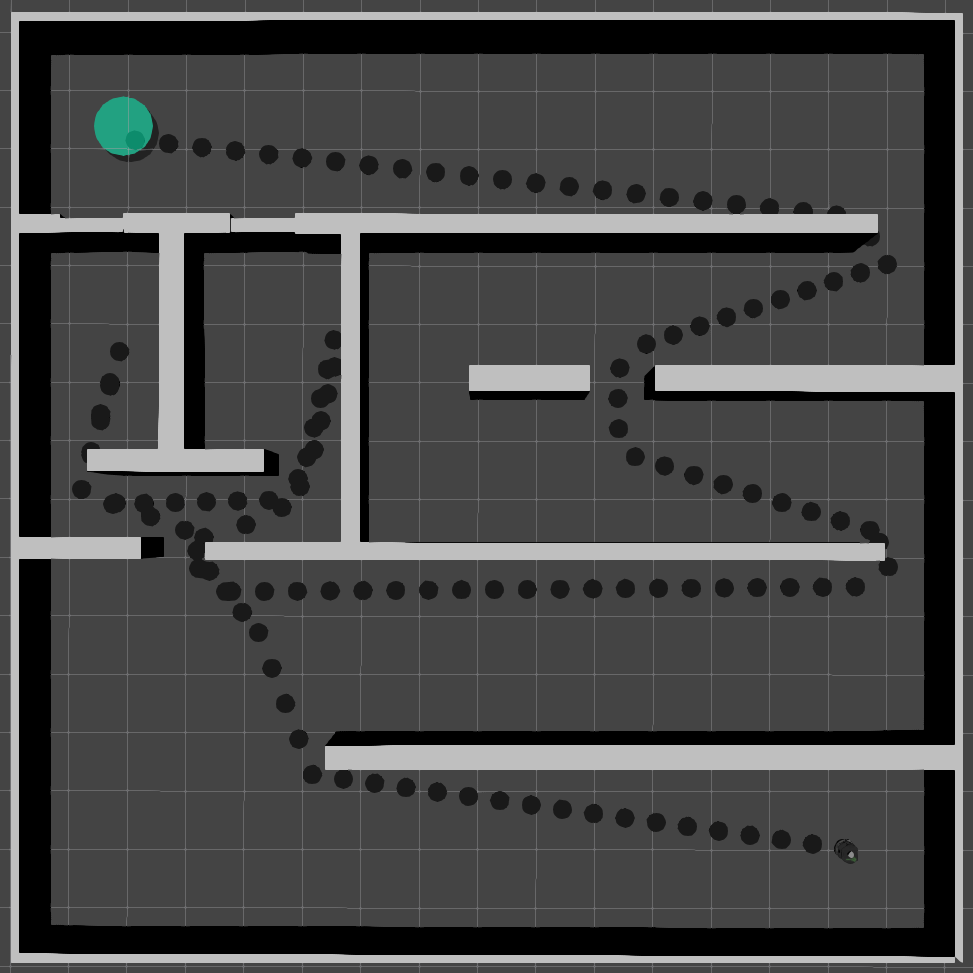}
         \label{fig:problem_b_world_4}
	\end{subfigure}\hfill	
	\begin{subfigure}[b]{0.33\linewidth}
		\centering
		\includegraphics[width=\textwidth]{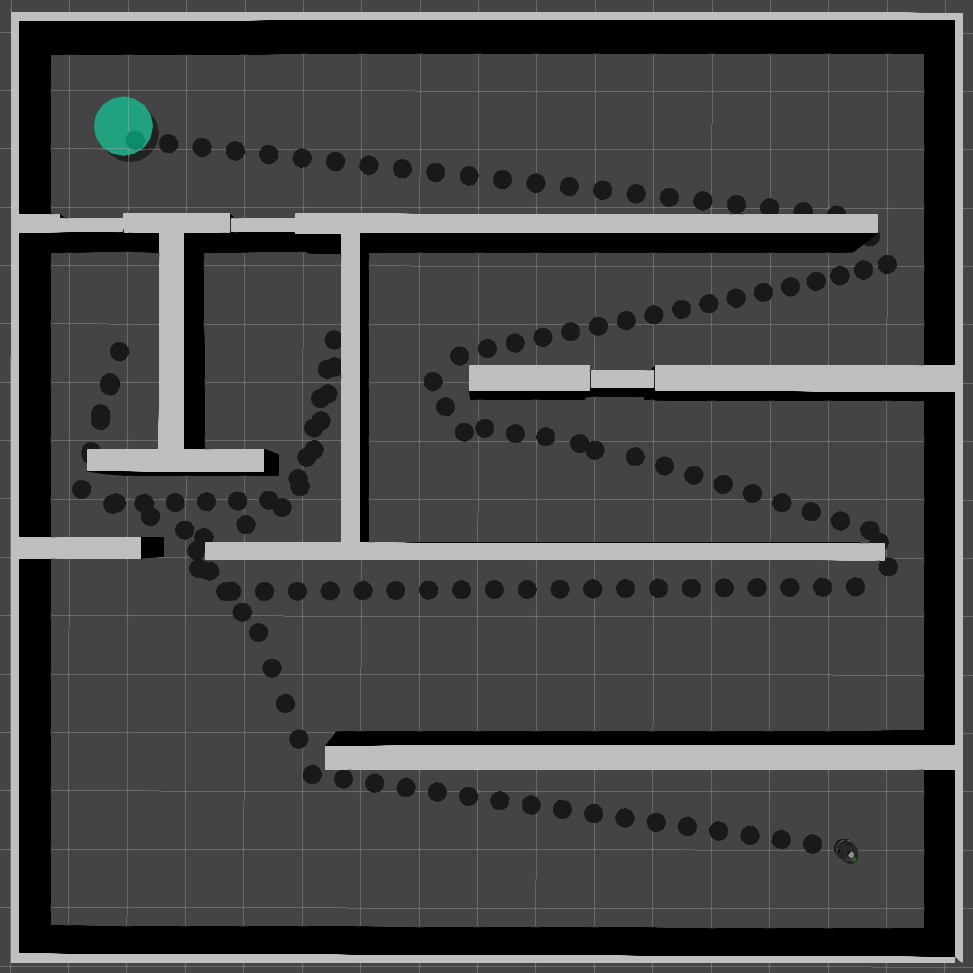}
         \label{fig:problem_b_world_5}
	\end{subfigure}\hfill
	\caption{Path-tree execution: The robot explores the doors on the left of the map. It potentially has to back-track and take the long route if it is blocked.}
	\label{fig:problem_b_path_executions}
\end{figure}

\setlength{\columnsep}{6pt}%
Fig.~\ref{fig:problem_b} shows the path-tree obtained for the problem-B, and Fig.~\ref{fig:problem_b_path_executions} shows its execution in a subset of the possible worlds. The higher number of doors leads to a path-tree which is much more complex.

Table~\ref{tab:planning_times_navigation}, gives the expected costs to goal, and the planning times obtained when running PTO 20 times. Each problem is solved using 2 different strategies for the RRG sampling: In the initial variation, the RRG sampling is stopped as soon as completeness is guaranteed (see \ref{section:rapidly-exploring-random-graph}), and the second variations (A$^\prime$ and B$^\prime$) use a minimum number of 5000 iterations to obtain a more qualitative path-tree.

\begin{table}
\begin{center}
\footnotesize
\addtolength{\tabcolsep}{-4pt}
\begin{tabular}{|l||l|l|l|l|l||l|l|}
\hline
                             & \thead{\# of\\ iter}  & \thead{random\\graph\\creation} & \thead{belief-\\space\\expansion} & \thead{policy\\ extraction} & \thead{partial\\shortcut} & \thead{path\\cost} & \thead{total\\planning\\time (ms)} \\ \hline
A & \thead{433\\(107)} & \thead{0.72\\(0.44)} & \thead{0.98\\(0.51)} & \thead{0.54\\(0.27)} & \thead{0.49\\(0.28)} & \thead{18.6\\(0.37)} & \thead{2.78\\(1.4)} \\ \hline
A$^{\prime}$ & \thead{5000\\(0)} & \thead{30.9\\(3.7)} & \thead{29.1\\(2.80)} & \thead{16.7\\(1.1)} & \thead{0.31\\(0.05)} & \thead{18.17\\(0.1)} & \thead{77.1\\(6.02)} \\ \hline
B & \thead{3956\\(626)} & \thead{44.7\\(14.3)} & \thead{345\\(102)} & \thead{173\\(60.8)} & \thead{3.25\\(1.07)} & \thead{45.1\\(1.74)} & \thead{567\\(174)} \\ \hline
B$^{\prime}$ & \thead{5000\\(0)} & \thead{61.5\\(6.49)} & \thead{538\\(32)} & \thead{270\\(22.9)} & \thead{2.95\\(0.38)} & \thead{43.9\\(1.25)} & \thead{873\\(47.1)} \\ \hline
\end{tabular}
\end{center}
\caption{Planning time and expected cost: the mean value is given first. The standard-deviation is in parentheses. Times are in milliseconds.}
\label{tab:planning_times_navigation}
\end{table}

Generally, the algorithm is fast, and takes less than 1~second even for the most complex map. The longest is the graph expansion to belief-space (see \ref{section:graph_expansion_belief_space}). The reason is that the number of belief-states grows quickly when the number of doors increases, and the observation model is queried for each belief-space node.

Increasing the number of iterations results in smaller expected costs of the path-tree, at the expense of the overall runtime. In practice, the final refinement step (see \ref{section:refinement}) leads to path-trees that are already near-optimal even without a very high number of iterations, such that it is an efficient strategy to keep a small number of minimal iterations.
\subsection{Robot arm object fetching} \label{experiment:mobile_manipulation}
\begin{figure}[!htb]
 \center{\includegraphics[width=0.45\textwidth]{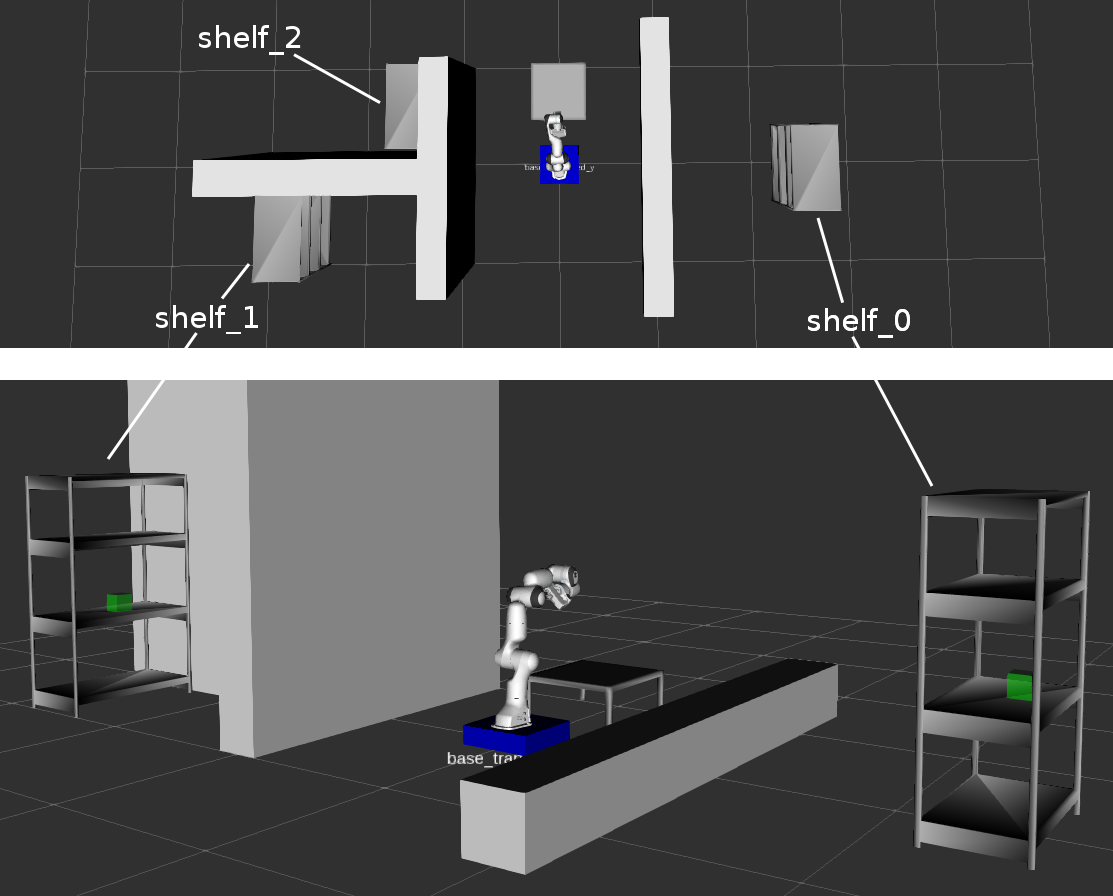}}
 \caption{Problem-A: The robot has to pick the green block. The block location is unknown, it may be on each one of the three shelves.}
 \label{fig:arm_example_1}
\end{figure}

\begin{figure}[!htb]
 \center{\includegraphics[width=0.45\textwidth]{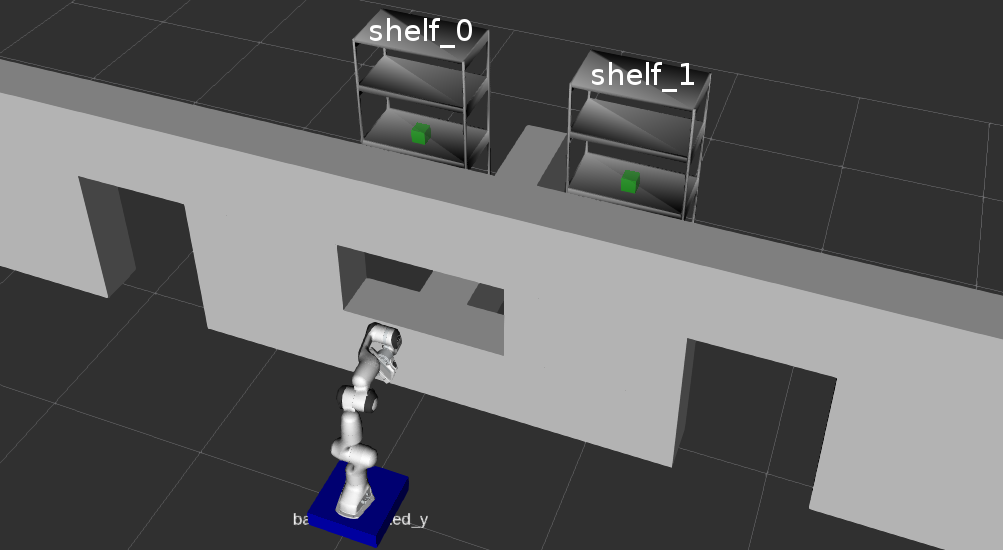}}
 \caption{Problem-B: The block might be on two different shelves. The opening between the two doors potentially allows the robot to observe the shelves without entering the rooms.}
 \label{fig:arm_example_2}
\end{figure}

Our next scenario is a Panda arm robot mounted on a mobile base. We introduce two examples, where the task is to pick-up an object which is at an unknown location (see problem-A on Fig.~\ref{fig:arm_example_1} and problem-B on Fig.~\ref{fig:arm_example_2}). The robot has prior knowledge of several potential locations, but the actual location is unknown at planning time. 

The observation model simulates that a sensor is placed on the robot gripper and that the perception pipeline detects the block when it is within the sensor field of view (60$^{\circ}$), at a distance less than 2 meters from the sensor, and not occluded by other objects (e.g. walls).

\begin{figure}[!htb]
 \center{\includegraphics[width=0.45\textwidth]{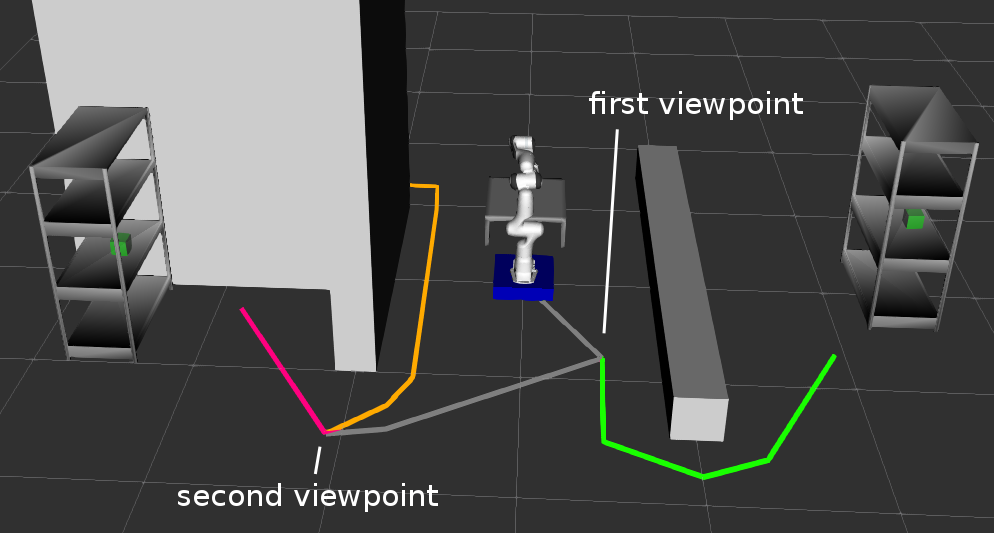}}
 \caption{Path-tree for Problem-A: The path tree first moves towards the shelf\_0 to reach the first viewpoint. If the object is on the shelf\_0, the green path is executed. Otherwise the robot moves towards the second viewpoint.}
 \label{fig:example_1_path_tree}
\end{figure}

Planning is performed in joint space with 9 degrees of freedom (2 for the base, and 7 for the robot arm). 
Fig.~\ref{fig:example_1_path_tree} shows the trajectory of the robot base. There are two observation branching points corresponding to the observations of the two shelves.

\begin{figure}[!htb]
 \center{\includegraphics[width=0.45\textwidth]{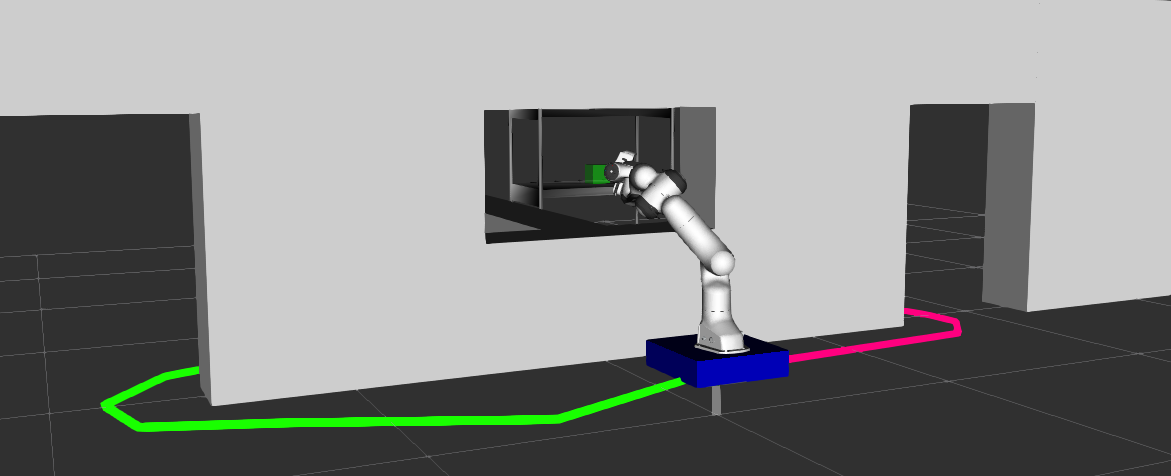}}
 \caption{Path-tree for Problem-B: A first common branch of the path-tree leads the robot to an observation point through the opening. Depending on the observation, the robot will execute the green or the magenta path.}
 \label{fig:example_2_view_point}
\end{figure}

Fig.~\ref{fig:example_2_view_point} shows the robot configuration at the branching point of the path-tree. The opening in the wall allows the robot to take a look at the shelf.

We report in Table~\ref{tab:planning_times} on the average path-tree cost and planning time obtained over 100 planning queries.
\begin{table}[!ht]
\begin{center}
\footnotesize
\addtolength{\tabcolsep}{-4pt}
\begin{tabular}{|l||l|l|l|l|l||l|l|}
\hline
                             & \thead{\# of\\ iter}  & \thead{random\\graph\\creation} & \thead{belief-\\space\\expansion} & \thead{policy\\ extraction} & \thead{partial\\shortcut} & \thead{path\\cost} & \thead{total\\planning\\time (s)} \\ \hline
A & \thead{10521\\(16481)} & \thead{5.86\\(14.7)} & \thead{0.31\\(0.66)} & \thead{0.30\\(0.66)} & \thead{0.43\\(0.06)} & \thead{3.27\\(0.53)} & \thead{6.75\\(15.8)} \\ \hline
B & \thead{4578\\(3484)} & \thead{1.28\\(2.05)} & \thead{0.03\\(0.04)} & \thead{0.01\\(0.02)} & \thead{0.26\\(0.03)} & \thead{3.72\\(0.46)} & \thead{1.59\\(2.10)} \\ \hline
\end{tabular}
\end{center}
\caption{Path-tree cost and planning time obtained over 100 planning queries: It indicates the mean value and the standard deviation in parentheses. Planning times are in seconds.}
\label{tab:planning_times}
\end{table}
In contrast to the examples of the previous section \ref{experiment:navigation}, the planning time is dominated by the random graph creation. This is consistent with the fact that the geometric collision checks are more computationally expensive with this robot. In addition, we note that the planning times have a large dispersion around the mean value (see the standard deviation on \ref{tab:planning_times}), which is due to the higher geometric complexity and dimensionality of the planning problem. Overall, the planning time remains limited to a few seconds.

\subsection{Comparison to baselines} \label{experiment:baseline comparison}
We compare PTO to two baselines on variations of the shelf domain (Fig.~\ref{fig:baseline_comparison}), where we plan for the 2D base position only. The number of shelves is varied to compare the scalability. 

In the first baseline, planning is hierarchical: it interleaves a symbolic Branch and Bound (B\&B) search \cite{MORRISON201679} defining high level plans, and a motion planning phase using RRT$^{*}$. 
The B\&B search explores the possible sequences for visiting the shelves. The nodes of the search correspond to shelves. For each node/shelf, 2 paths are planned using RRT$^{*}$: a path to an observation point, and a path to fetch the object (to be executed if the object is actually detected). The path-tree is obtained by gathering the path pieces from the root node to a leaf of the search tree.  
The B\&B search is performed in a depth-first fashion, which leads to a quick first solution. The best path-tree found so far is an upper bound of the cost and allows the pruning of a majority of the search tree.

The second baseline is adapted from the Orbital-Bellman-Tree (OBT) method \cite{vega2020asymptotically}\cite{mm-prm}. The algorithm is adapted to belief-space planning by considering that different belief states correspond to different orbits, and by using Algorithm~\ref{alg:expected_costs} to perform contingent planning and retrace a path-tree (instead of extracting a simple path). It grows different roadmaps for each visited belief state using PRM$^{*}$. Observation points are explicitly sampled which transition between roadmaps. 

We report on Figure \ref{fig:costs_and_runtime} on the average expected cost of the path-trees, as well as on the planning times obtained over 100 planning queries. RRT$^{*}$ paths are planned with 2500 iterations. The random graph of PTO is created with a minimum number of 5000 iterations. OBT+PRM$^{*}$ is parameterized to reach an average of 5000 iterations per roadmap.

\pgfplotsset{width=5cm, height=7cm}
\begin{figure}[h]  
 \centering
  \begin{subfigure}[t]{0.45\linewidth}
   \center{\includegraphics[width=\textwidth]{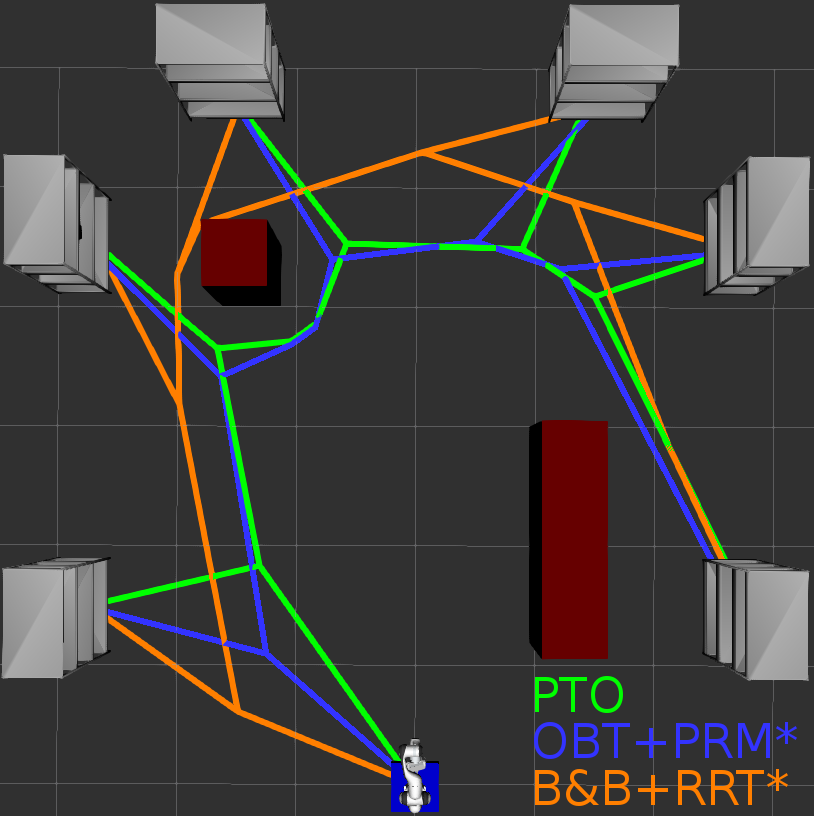}}
   \caption{PTO plans path-trees with lower costs than RRT$^{*}$ and PRM$^{*}$.}
   \label{fig:baseline_comparison}
  \end{subfigure}
  \begin{subfigure}[t]{0.26\linewidth}
    \begin{tikzpicture}[scale=0.55]
	 \begin{axis} [
	    ybar = .05cm,
    	bar width = 3pt,
    	ymin = 0,
    	ymax = 3,
    	xtick = data,
    	enlarge y limits = {value = 0.05, upper},
    	enlarge x limits = {abs = .8},
    	legend pos=north west
		]
		\addplot [green!20!black,fill=green!80!white] coordinates {
		(2, 1.7529471182990233) (4, 2.2212974330517827) (6, 2.3911496782081954) (8, 2.360321441477037)
		};
		\addplot [blue!20!black,fill=blue!80!white] coordinates {
		(2, 1.84) (4, 2.245) (6, 2.441) (8, 2.485)
		};
		\addplot [orange!20!black,fill=orange!80!white] coordinates {
		(2, 1.8569080235157531) (4, 2.3748431227440063) (6, 2.6112924196096783) (8, 2.5971396786963075)
		};
		\end{axis}
	\end{tikzpicture}
    \caption{Path cost (m)}
    \label{fig:cost}
  \end{subfigure}
  \begin{subfigure}[t]{0.26\linewidth}
    \begin{tikzpicture}[scale=0.55]
	 \begin{axis} [ 
	    ybar = .05cm,
	    ymin = 0.05,
    	bar width = 3pt,
    	ymode=log,
    	log origin=infty,
    	xtick = data,
    	enlarge y limits = {value = .1, upper},
    	enlarge x limits = {abs = .8},
    	legend pos=north west
		]
		\addplot [green!20!black,fill=green!80!white] coordinates {
		(2, 0.0838456953) (4, 0.3447605707) (6, 1.3742766512) (8, 7.585573182200001)
		};
		\addplot [blue!20!black,fill=blue!80!white] coordinates {
		(2, 0.262) (4, 1.674) (6, 6.847) (8, 33.91)
		};
		\addplot [orange!20!black,fill=orange!80!white] coordinates {
		(2, 0.0763110857) (4, 0.8065830762000001) (6, 11.0703240974) (8, 107.3652537214)
		};
		\legend {PTO, OBT+PRM$^{*}$, B\&B+RRT$^{*}$}; 
		\end{axis}
	 \end{tikzpicture}
     \caption{Planning time (s) in logarithmic scale}
     \label{fig:time}
   \end{subfigure}
   \caption{Comparison of PTO to baseline algorithms: \ref{fig:cost} and \ref{fig:time} give the costs and planning times (log scale) for different number of shelves. PTO provides path-trees with lower costs and scales better.}
   \label{fig:costs_and_runtime}
   \vspace{-1.5em}
\end{figure}

First, we observe that PTO consistently leads to the lowest path-tree costs. This can be understood compared to B\&B+RRT$^{*}$: the decomposition into piecewise motions leads to path pieces that are optimal w.r.t the sub-problem of reaching one shelf, but taken together, they don't form an optimal path-tree. In contrast, PTO searches for a globally optimal path-tree. This is visible on Fig.~\ref{fig:baseline_comparison}: the B\&B+RRT$^{*}$ path-tree greedily moves towards the next shelf to explore, whereas PTO takes a less direct path to the observation point, which overall leads to a shorter path-tree. 
PTO also leads to lower costs than OBT+PRM$^{*}$. Even though OBT+PRM$^{*}$ is also asymptotically optimal (see~\cite{mm-prm}), we observed, that for a comparable final number of nodes, the belief-graph of PTO (grown as a RRG with periodic sampling of goal states) is more efficiently sampled in the regions of the optimal path-tree compared to the goal-agnostic PRM$^{*}$ graph.

Second, PTO scales better w.r.t. the number of shelves (see Fig.~\ref{fig:costs_and_runtime}), it outperforms B\&B+RRT$^{*}$ by one order of magnitude on the 8-shelves problem.
The main reason is that PTO is more sample-efficient: the random graph creation which involves state and transition checks is constructed just once. On the other hand, the B\&B+RRT$^{*}$ samples a new random tree for the planning of each path-piece. OBT+PRM$^{*}$ is more efficient but still requires, in general, different roadmaps for each belief-state to account for domains where the partially observable state impacts the state and transition validities (like for the problem of Fig.~\ref{fig:intro}).

In this example, the three methods could be made faster by using, e.g. heuristics for the B\&B tree search or by re-using samples across belief-states for the OBT+PRM$^{*}$. We chose, however, to compare the general approaches, without optimizations tailored to the particular example.

\section{Conclusions}
We proposed a new sampling-based path planning algorithm (PTO) for motion planning problems where
discrete and critical aspects of the world are partially observable. The algorithm not only optimizes feasible motions but also plans observation points that allow the robot to gain knowledge about its environment to achieve its task. The resulting motions are path-trees in belief-space that react to observations. We showed that PTO is asymptotically optimal, and it compares advantageously to Task and Motion Planning (TAMP) approaches both in terms of optimality and runtime efficiency.

\bibliography{references_b}
\bibliographystyle{ieeetr}

\end{document}